\documentclass[conference]{IEEEtran}
\newcommand\Mark[1]{\textsuperscript{#1}}

\IEEEoverridecommandlockouts
\usepackage{cite}
\usepackage{amsmath,amssymb,amsfonts}
\usepackage{algorithmic}
\usepackage{graphicx}
\usepackage{textcomp}
\usepackage{xcolor}
\def\BibTeX{{\rm B\kern-.05em{\sc i\kern-.025em b}\kern-.08em
    T\kern-.1667em\lower.7ex\hbox{E}\kern-.125emX}}
\begin{document}

%

\title{Predicting Real-time Scientific Experiments Using Transformer models and Reinforcement Learning\\
	
	{\normalfont\large 
		Juan Manuel Parrilla-Gutierrez\Mark{1,2}
	}\\[-1.5ex]

\thanks{To citate please use the copy in: 2021 20th IEEE International Conference on Machine Learning and Applications (ICMLA), doi: 10.1109/ICMLA52953.2021.00084. The research described here was developed while working at the University of Glasgow. This arxiv manuscript was prepared in my current organisation, Glasgow Caledonian University. If you need to contact me, please use the e-mail address provided above.} 
}

\author{
	\IEEEauthorblockA{%
		\Mark{1}School of Chemistry, University of Glasgow\\
		Glasgow, United Kingdom%
	}
	\and
	\IEEEauthorblockA{%
		\Mark{2}School of Engineering, Glasgow Caledonian University\\
		Glasgow, United Kingdom\\
		juanma.parrilla@gcu.ac.uk
	}
}

\maketitle

\begin{abstract}
Life and physical sciences have always been quick to adopt the latest advances in machine learning to accelerate scientific discovery.
Examples of this are cell segmentation or cancer detection.
Nevertheless, these exceptional results are based on mining previously created datasets to discover patterns or trends.
Recent advances in AI have been demonstrated in real-time scenarios like self-driving cars or playing video games.
However, these new techniques have not seen widespread adoption in life or physical sciences because experimentation can be slow.
To tackle this limitation, this work aims to adapt generative learning algorithms to model scientific experiments and accelerate their discovery using \textit{in-silico} simulations.
We particularly focused on real-time experiments, aiming to model how they react to user inputs.
To achieve this, here we present an encoder-decoder architecture based on the Transformer model to simulate real-time scientific experimentation, predict its future behaviour and manipulate it on a step-by-step basis.
As a proof of concept, this architecture was trained to map a set of mechanical inputs to the oscillations generated by a chemical reaction. 
The model was paired with a Reinforcement Learning controller to show how the simulated chemistry can be manipulated in real-time towards user-defined behaviours.
Our results demonstrate how generative learning can model real-time scientific experimentation to track how it changes through time as the user manipulates it, and how the trained models can be paired with optimisation algorithms to discover new phenomena beyond the physical limitations of lab experimentation.
This work paves the way towards building surrogate systems where physical experimentation interacts with machine learning on a step-by-step basis.
\end{abstract}


\section{Introduction}

Scientific experimentation can generate large datasets, but these are often difficult to analyse because they might contain inconsistent data \cite{Brown2018}.
Machine learning can alleviate some of these problems.
Today, deep learning is used to detect cancers from biomarkers \cite{Ardila2019}, or achieve human-like results in problems such as cell segmentation \cite{Moen2019}.
However, these results are often based on mining datasets that were previously collected, and real-time AI techniques \cite{Bojarski2016} have not seen widespread adoption because data generation can be time intensive.

\begin{figure*}
	\centering
	\includegraphics[width=0.95\textwidth]{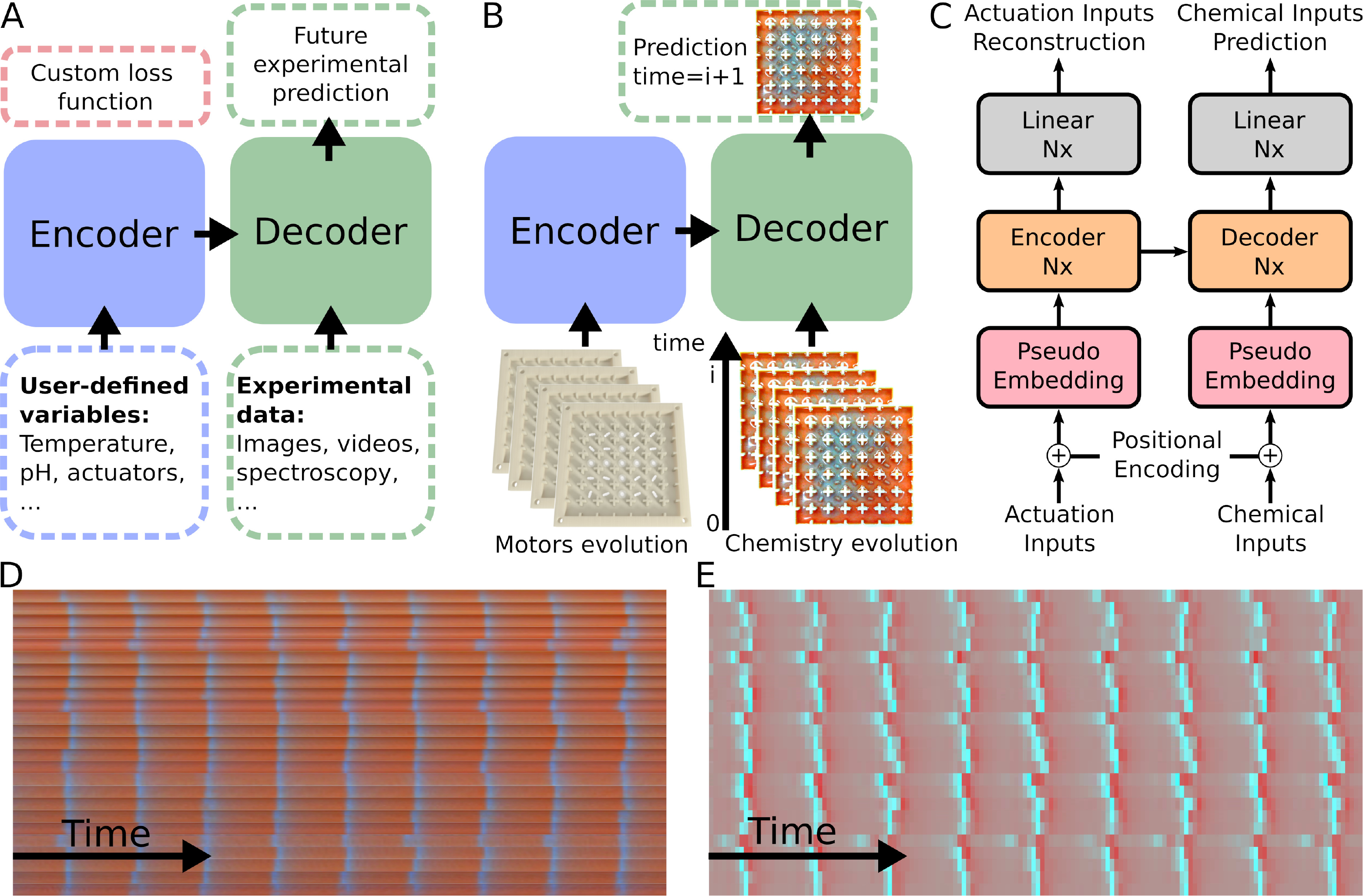}
	\caption{A: This work proposes to use encoder-decoder architectures with real-time experimental data. B: We focused on a chemical reaction that produces oscillations when stirred. C: Our architecture used the Transformer model. D: Chemical oscillations as generated by a physical platform and (E) as generated by our model.}
	\label{F1}
\end{figure*}

To increase experimental data throughput, this work aims to simulate scientific experimentation using generative learning.
To achieve this, we focused on using an encoder-decoder architecture, where the encoder is used to process user-defined variables, and the decoder to model the actual experiment.
This way, our target is to model how user inputs (through the encoder) impact real-time experiments (as outputted by the decoder, Fig. \ref{F1}-A).
In particular, we focused on modelling an oscillatory chemical reaction placed in a five-by-five array of weakly connected cells, and how the oscillations within each cell could be controlled using magnetic stirrers, see Fig. \ref{F1}-B.
In this experiment the relation between magnetic stirrers and chemistry is not one to one, and there is no known model that can fully describe it.
Therefore, compared with similar problems where the physical phenomena are modelled based on known systems, here we aimed to model it using only experimental data.

This work focuses on the Transformer model (Fig. \ref{F1}-C), among the several encoder-decoder architectures, since it is currently outputting very promising results in translation tasks \cite{Vaswani2017}.
The original model, which from now on will be referred as ``vanilla Transformer'', aimed at solving language translation problems with an encoder-decoder architecture that used ``attention'' layers.
This work shows how the Transformer model can be further modified to work with  scientific data comprising of both user-defined variables, such as temperature or stirring patterns, and raw experimental data, such as the video of an on-going experiment.
Finally, we focused on using the trained model with optimization or exploratory algorithms, such as Evolutionary Algorithms or Reinforcement Learning, to discover novel phenomena.
The specific contributions of this work are:

\begin{itemize}
	\item A Transformer architecture to model and predict how a real-time scientific experiment changes over time based on user-defined inputs.
	\item A Reinforcement Learning controller that consists of a single layer of neurons and uses the attention weights as returned by the Transformer's decoder to calculate its reward function.
	\item A demonstration of how the Transformer model can act as a kernel-like function to increase the feature space.
\end{itemize}

\section{Related Work}

Generative modelling aims to describe how a dataset is generated, in terms of a probabilistic model. 
By sampling from this model, it can generate new data \cite{Goodfellow2020}.
This technique, paired with Deep Learning, has been used to teach machines how to paint \cite{Gatys2016} or compose music \cite{Dong2018}.
This work aims to use generative modelling to model scientific experiments for which no known theoretical model exists, but only experimental data.
As an example of this kind of experiment, we focused on modelling the behaviour of a chemical oscillator, which is a type of chemical reaction that continuously oscillates between two or more states \cite{Epstein1998}.
In particular, we focused on modelling how the oscillations can be manipulated via user input \cite{Dutt1993}.
This type of experiment has been modelled using theoretical models, such as the Oregonator model \cite{Hsu1994}.
However, this model is a simplistic approximation and it fails to capture the dynamics of real world complex experiments, like the chemical oscillator described in \cite{Parrilla-Gutierrez2020}.

To achieve our aims, we focused on encoder-decoder architectures that are commonly used in natural language processing \cite{Wang2017a}.
Among the different encoder-decoder architectures, we focused on the Transformer model \cite{Vaswani2017}.
Therefore, the main technical objective of this work was to adapt the Transformer model to work with experimental data.
This objective is similar to recent work in the field of physics where machine learning has been used to model physical phenomena from known models using graph networks \cite{Sanchez-Gonzalez2020}, encoder-decoder recurrent neural networks (RNN) \cite{Wiewel2019,Colen2021} or Generative Adversarial Networks \cite{Xie2018}.
As opposed to physics, chemistry rarely has known theoretical models, and chemists usually work using their own intuition and empirical evidence. 
Based on this, chemists have developed the field of ``chemoinformatics'' where machine learning has been used extensively \cite{Lo2018,Sterling2021}, and recently generative architectures similar to the ones developed in this work have also been used \cite{Gao2020,Bort2021,Zhou2017}.
The main difference in this work is that we did not aim to predict synthesizability, but we focused instead on tracking real-time experimental changes based on user-input.
AI has also been used to guide experimentation \cite{Duros2017, Parrilla-Gutierrez2014}.
Nevertheless, in these publications machine learning was used to propose experiments that were tested in the real world, while this work aimed to simulate \textit{in-silico} the experiment.

\section{Data collection and processing}
In related work \cite{Parrilla-Gutierrez2020} a platform was built to manipulate a chemical oscillator in a five-by-five array of weakly connected cells. 
Each cell contained a magnetic stirrer that was rotated using a motor. 
These rotations stirred the chemistry, that eventually generated oscillations that were related to the configuration of motors used: which motors were enabled, their directions, and their speeds. 
Using the described physical platform, a series experiments were performed where different combinations of motors were activated at different times. 
The objective was to study both how a given motor configuration gave rise to different chemical oscillations, and how the oscillatory nature of the chemical reaction would impact newer motor configurations and therefore new chemical oscillations, as oscillations can continue to be observed even after stirring is disabled. 
For each experiment we saved a video of the experiment, and the motor configuration used at every time-step. Finally, the data was split into sequences.

\section{Results}

\subsection{Using the Transformer model to predict experiments}

Compared to the ``vanilla Transformer'', the four main differences in this work are (Fig. \ref{F1}-C):
(1) The decoder softmax layer was replaced with a dense layer with same size as the decoder's input feature space.
This way, it could solve regression problems as the one we were targeting.
(2) The embedding layers were replaced with dense layers.
(3) We added a new output to the encoder that aimed to reconstruct its input.
This reconstruction was added because very often lab experiments have some degree of inertia, meaning that there is a lag between user inputs and how the experiment behaves.
We found that without forcing the encoder reconstruction as part of the learning process, the decoder would ignore the encoder (ie. user defined variables), and it predicted the next state of the experiment using only the previous states of the experiment (the decoder input).
(4) Both the encoder and the whole Transformer were trained in cycles, where initially the encoder was trained for a few iterations, and then the whole Transformer, including the encoder, was trained for more iterations. 

\begin{figure*}
	\centering
	\includegraphics[width=0.75\linewidth]{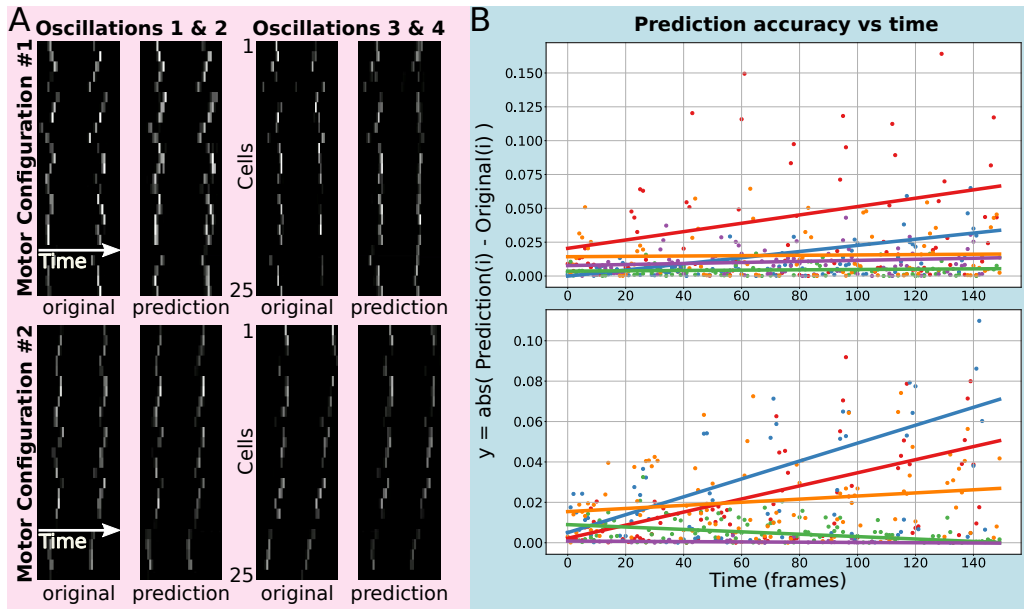}
	\caption{A: Comparing the first four oscillations -- from two different motor configurations -- as generated using the Transformer model with data from the test set. B: Comparing the prediction generated using the Transformer model with data from the test set over a full experiment. Each plot shows five different experiments. The oscillations at time $i$ in the original (test) dataset were compared with the oscillations in the prediction at time $i\pm{}w$ to account for phase differences. The error represented in the $y$ axis is per cell.}
	\label{F2}
\end{figure*}

Using the described architecture, our Transformer model was trained to learn associate sequences of motor actuations to sequences of experimental data, and then use this association to predict subsequent behaviour.
The Transformer's attention mechanism was critical for the success of this task, because the experimental data represented oscillations that might appear for one or two seconds, with a periodicity of 40 to 60 seconds, meaning that for more than 95\% of the time, the signal to train for was zero.
Attention layers can remember that there was an oscillation back in time, while other models, such as stacked LSTMs without attention mechanisms, failed to capture these dynamics.
Fig. \ref{F2} shows some of the Transformer predictions compared with the original data as generated by the physical platform.
Initially, the Transformer replicated the original data with a very high degree of accuracy, but as the experiment evolved, prediction and real experimental data begin to diverge.
Nevertheless, the Transformer still produced outputs that resembled an oscillatory chemical reaction, showing that the described model can generate data similar to physical platform. 
The advantage of this model resides on its data throughput.
This enabled us to pair the simulated platform with different optimization or exploratory algorithms to discover new phenomena, in a way that was not possible before.

\subsection{Pairing the Transformer model with a Genetic Algorithm}

\begin{figure*}
	\centering
	\includegraphics[width=0.99\linewidth]{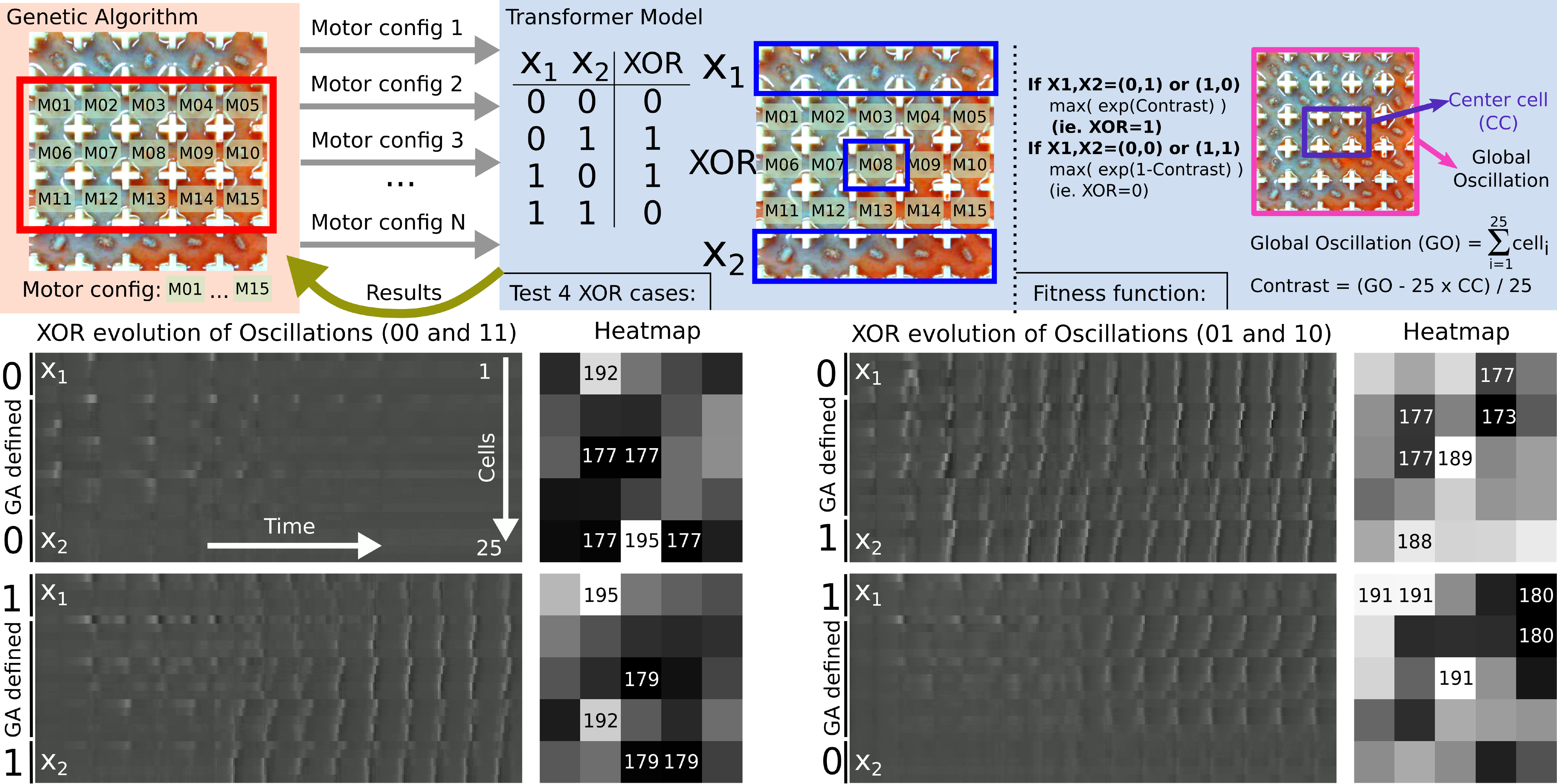}
	\caption{A GA was used to find motor configurations that would activate the chemistry to oscillate like an XOR gate. The first and fifth row were used as inputs, while the centre cell was used as output. These motor oscillations would then be tested using the trained Transformer model. The bottom side of the figure shows a particular motor configuration and how it emulates a XOR. The accumulated result of the oscillations can be seen on the heatmaps.}
	\label{F4}
\end{figure*}

Herein, we paired the model with a Genetic Algorithm (GA) to find configurations of motors that manipulated the chemistry to behave in a user-defined way.
In particular, the GA was used to find motor configurations where the chemistry behaved like an XOR gate.
To accomplish, the first and fifth rows of the 5 by 5 array were defined as the XOR inputs, and centre cell as the XOR output (Fig. \ref{F4}).
The inputs could be set to either 0, meaning the motors did not rotate, or 1, they rotated at full speed.
The output was considered 1 if the centre cell oscillated more, on average, than the other 24 cells, and it was considered 0 if it oscillated less.
Here by ``oscillation'' we meant the cell value as outputted by the Transformer -- which was trained using blue channel data.
Therefore, higher values mean the blue colour was stronger, that is associated with an oscillation, while no oscillations is associated with red colour.
Following this, the GA aimed to find motor configurations for the motors in rows two, three and four, whereby enabling or disabling the first and fifth row, the output (centre cell) behaved like an XOR for its four cases.
That is, the centre cell would oscillate less than the other cells if the first and fifth row were both either enabled (11) or disabled (00), and it would oscillate more if one of them was enabled and the other disabled (01 and 10), see Fig. \ref{F4}.

\subsection{Pairing the model with Reinforcement Learning}
\label{RL}

\begin{figure*}
	\centering
	\includegraphics[width=0.75\linewidth]{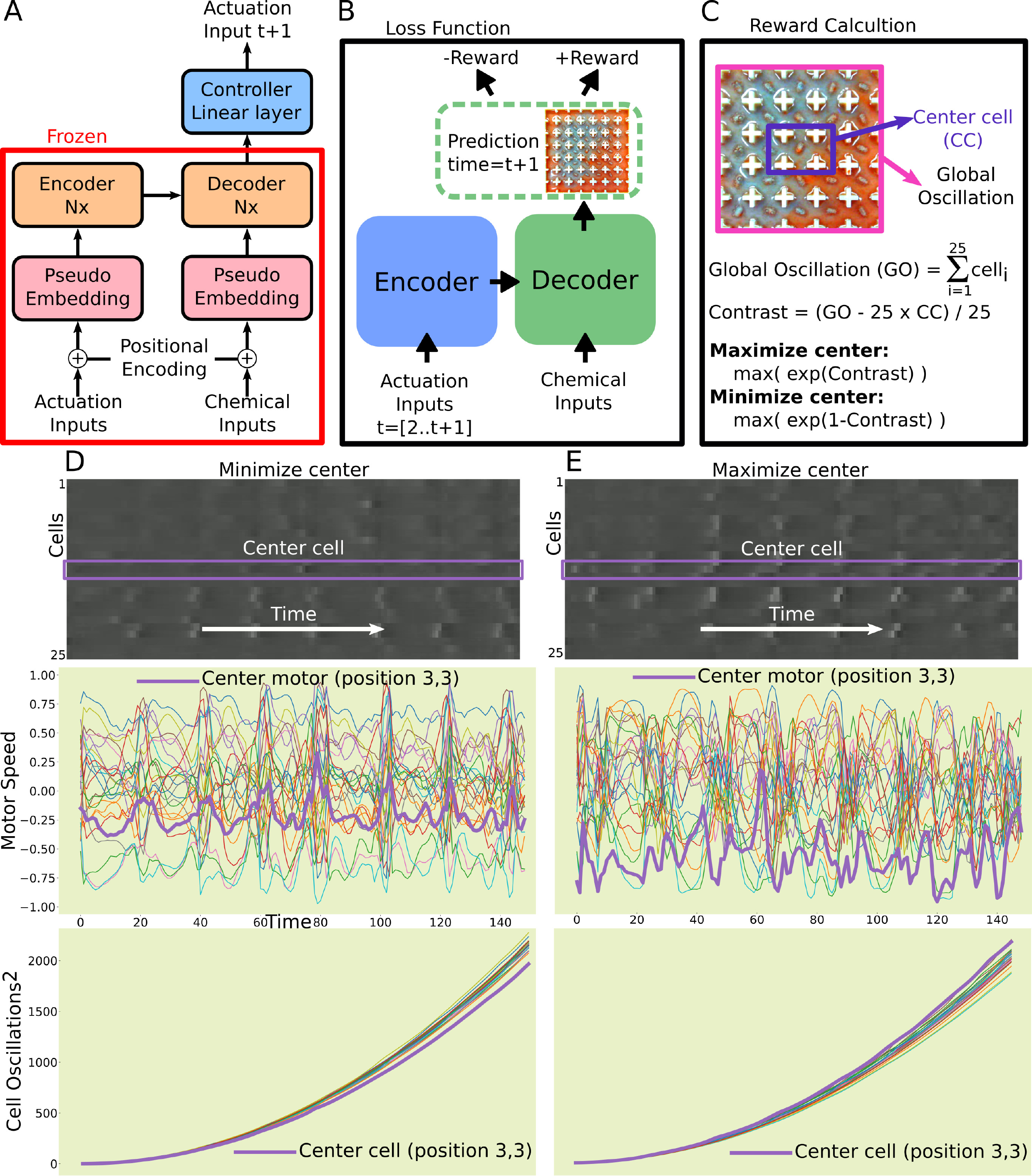}
	\caption{A: To implement Reinforcement Learning, we connected the output of the encoder to a single layer of neurons that acted as controller. B: The loss function tested the motor configuration as provided by the controller using the trained Transformer model. C: The motor configuration was given a positive reward if it increased (or decreased) the oscillations of the centre cell. D: Example of using the controller to minimise the centre cell. E: Similar example but using the controller to maximise the centre cell. In both D and E, a motor speed of 0 means the motor is not moving, a motor speed of 1 means full speed clockwise, while a speed of -1 means full speed counter-clockwise. The two bottom plots named ``Cell oscillations$^2$'' describe the blue channel of each cell as the experiment progressed. It was squared for better visualization.}
	\label{F5}
\end{figure*}

One of the main features of the original platform is that different motor speeds can be applied at any time. 
In this kind of problems, where the ``action'' can change as the problem evolves, Reinforcement Learning (RL) can offer better results than a GA.
Following the RL implementation from \cite{Ha2018}, our controller was a single layer of neurons that was directly connected to the output of the decoder (Fig. \ref{F5}-A).
This very simple controller took advantage of the fact that the Transformer model had already been trained to predict how the data would behave.
In particular, our controller was trained to output the action (i.e. the motor configuration) needed to ``push'' the chemistry towards a user-defined behaviour.

To evaluate the quality of a motor configuration outputted by the controller, the new motor configuration was added to the ``actuation inputs'' that were used on the forward pass to output the given motor configuration, and these new actuation inputs plus the old chemical inputs were feed forward the transformer model described in Fig. \ref{F1}-C.
This would output a new chemical prediction that would reflect how the new motor configuration impacted the behaviour of the chemistry (Fig. \ref{F5}-B).
The task to solve by the ``agent'' was be to find motor configurations that maximised or minimised the value of the centre cell against the other 24 cells in the array (Fig. \ref{F5}-C).
Because the value of a cell is related to its blue channel, and higher values of blue channel represent oscillations, then the aim was to keep the centre cell oscillations above (or below) average.

Fig. \ref{F5}-D shows an example where the centre cell was minimised.
It can be seen in the greyscale plot of cells oscillation vs time how it managed to suppress most oscillations.
Below this plot the evolution of the motor speeds through time is shown, with a focus on the motor in position 3,3, which is the centre cell itself. 
Here we see how the controller was continually tweaking the value of the motors. 
Interestingly, the motor speeds oscillated in a similar way to the chemical reaction, with sudden changes when the cell actually oscillated.
The bottom plot shows how the value of the centre cell was lower, on average, than all the others.
Fig. \ref{F5}-E is similar to the D panel, but focuses instead on maximizing the centre cell.
Both D and E started from the same state: Given the same chemical state, the controller managed to either maximize or minimize the objective function.

\subsection{Using the model to increase the experimental feature space}

\begin{figure*}
	\centering
	\includegraphics[width=0.75\linewidth]{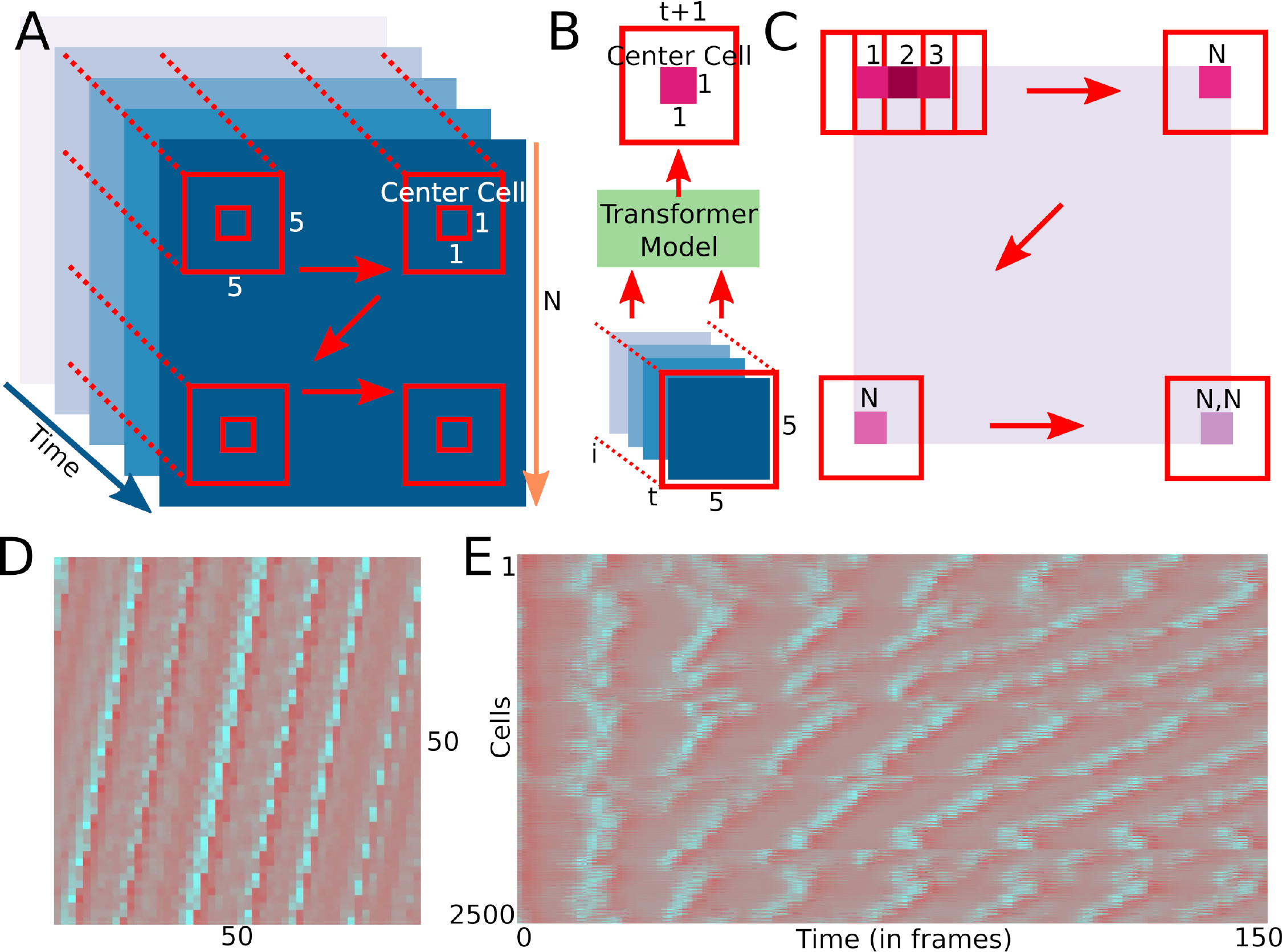}
	\caption{A: Different slices of time oscillations are taken, and inputted into (B) the trained transformer model, that acts similarly to a kernel. C: From the output of the transformer, only the centre cell is kept, and the results are stitched together to generate an output of size N by N. D and E: Following this, oscillations of size 25 by 25 were created.}
	\label{F6}
\end{figure*}

One of the main limitations of physical experimentation is the size of the feature space used to represent both input and output variables.
An example of this limitation can be seen on the platform used to collect the data used in this research: the complexity of chemical phenomena that can occur in such a small feature space (5 by 5 oscillation outputs) was limited.
To increase the feature space, our Transformer model, which had 5-by-5 sequences of data as the I/O, was used as pseudo-kernel function to generate outputs of size N-by-N based on inputs of size N-by-N (Fig. \ref{F6}-A).
To achieve this, the Transformer model was convolved along the N-by-N data inputs, in a similar way to Convolutional Neural Networks (CNN).
While CNNs convolve along multiple channels, our model convolved along multiple time slices, and where CNNs generate a convolution output by performing arithmetic operations between the kernel and the input data, our model generated an output by feed-forwarding the input data through the Transformer model (Fig. \ref{F6}-B and C).
Using this approach, we were able to generate an output space of, for example, size 50 by 50, which used a feature space of the same size as input (Fig. \ref{F6}-D-E). 
An interesting observation was how the simulation generated several ``oscillatory fronts'', that are very common when working with this chemistry, but that we never encountered in the 5-by-5 platform.

\subsection{Limitations and discussion}

The main limitation of the work is that while the quality of our modified Transformer model was validated against the test dataset, the quality of the results obtained using the GA and RL were not validated in a real platform, but the validation of their results was tested \textit{in-silico}.
The dataset we chose to test our approach (a chemical oscillator) was suitable due to being a small, while offering distinctive properties.
Nevertheless, this type of chemistry has a high degree of stochasticity.
Thus, it is unlikely that the results presented here could be transplanted to a real platform.
It is our objective in future research to choose a more deterministic system, which still doesn't have a theoretical models, and fully validate the results.

\section{Conclusion and Future Work}

In this work we presented a modified Transformer architecture to handle real-time scientific experiments. 
To do so, we explored how the Transformer's encoder could be used to interpret user-defined variables, while the decoder was used to simulate a scientific experiments.
It is, to our knowledge, the first purely generative model applied to real-time scientific experimentation, with a focus on generating real-time sequential data.
We showed how this model can interact with optimization or exploratory algorithms, such as Genetic Algorithms or Reinforcement Learning, and discover new phenomena in a way that was not be possible before due to the limitations of physical experimentation

We are excited about the future of generative learning applied to scientific experimentation beyond the current focus on class predictability. 
In particular when combined with automated platforms that can manipulate experiments in real-time. 
It is our future objective to apply the results described in this work on fully surrogate systems, where a robot would perform a physical experiment while the generative models perform high throughput \textit{in-silico} experimentation that is continuously validated against the real experiments.

\section{Methods and materials}

\subsection{Data preparation}
\label{dataprep}
Using the physical platform described, 55 different experiments were performed.
This platform contained an array of 5 by 5 weakly connected cells, where the fluid and chemistry could move through them.
Therefore, oscillations appearing in a cell could move to neighbouring cells. 
Thus, the whole array could produce global oscillations
Each cell contained a magnetic stirrer, and below each cell there was a DC motor with a magnet attached to its shaft. 
When any of the motors were actuated, the stirrer above them would also rotate, and thus stir the chemistry, which would eventually generate oscillations.
Each experiment lasted 30 minutes where different motors were enabled at different speeds during sequences of one or two minutes. 
Thus, each experiment tested 15 or 30 motor combinations. 
To perform these experiments, the motor configurations to be tested were chosen either randomly or following user-defined shapes. 
In the random experiments the motors were assigned randomly different speeds and directions in sets than went from sequences where only three motors were enabled, to sequences where all of them were.  
In the user-defined experiments, we defined simple shapes, like a column of motors enabled while all the other ones were disabled.

Once all the experiments were performed, the data was stored as tuples with as many entries as frames in the video, where one element was the input motor patterns, and the other element was the associated chemical state, as recorded from a camera placed in top of the platform.
The videos, which represented the evolution of the chemistry through the experiment, were processed in two different ways: 
(1) They were binarised into arrays with 25 cells of 0s and 1s. A 0 was assigned when no oscillation happened in that cell, while a 1 was assigned when an oscillation happened.
The cell binarisation based on the colour of the chemistry was performed using a Super Vector Machine.
(2) For every frame, which was recorded using the RGB colour-scheme, the blue channel was isolated. 
Then, a moving average of the blue value was calculated for each of the 25 cells, a it was subtracted from the average value to centre the signal around zero and dampen the color changes of the chemical reaction as the experiment progresses, while maintaining the oscillations. 
Finally, it was normalized between 0 and 1.

The data related to the motors configurations was normalized between -1 (max speed counter-clockwise) and 1 (max speed clockwise), being 0 no speed (motor disabled).
Before processing, every video was sped up by a factor of five, outputting 7200 frames.
Then, we generated sequences of 150 elements, where each sequenced sampled the 7200 frames every 8 frames, with a stride of one.
This means that for example, the very first sequence fetched frames: ``0, 7, 15, \ldots '' up to having 150 elements, while the second sequence fetched frames: ``1, 8, 16, \ldots''.
The last step converted the overall data into batches of size 64.
This step was done using Keras' ``TimeseriesGenerator''.
The batch size used was 64, which is the maximum we could use with our GPU without having memory problems.
Therefore, the final data had a shape like (64,150,25) for both the motor configurations and the chemistry evolution.

\subsection{Transformer architecture and training}
\label{transarch}

Architecture: Dense layers were used instead of embedding layers. 
The input data was already in the format (seq\_length, input\_features).
Thus, these dense layer transformed it into (seq\_length, latent\_space\_length).
We obtained the best results using dense layers of size 128.
The last ``softmax'' layer in the ``vanilla Transformer'' was replaced with a ``relu'' dense layer -- or a ``sigmoid'' one in case of using the binarized dataset.
We obtained the best results with a dense layer of size 1024 between the output of the last decoder layer and the final output dense layer -- which was of size 25 since that's the size of our experimental feature space.
The encoder output had two similar extra dense layers, one with size 1024, and the output one with an ``elu'' activation function.
Finally, in our implementation we obtained the best results with 4 stacked encoders and decoders (the ``vanilla Transformer'' uses 6).

Training: 
Following the ``vanilla Transformer'' we used a Custom Scheduler with a variable learning rate with 5000 warm-up steps.
We used a dropout rate of 0.2.
The model was trained with Adam optimizer, same parameters are ``vanilla Transformer''.
The loss function used was a slightly modified ``mean squared error'' (mse). Ror every time-step in the sequence we calculated its mse, and also the maximum element from ``y\_true''. 
Then we divided the time-step related mse with the time-step related max.
Finally, the average per time-step was returned for the whole sequence
We applied this small scaling factor to the mse because our dataset represents a chemical oscillator which has a value of 0 most the time, and we wanted the learning to particularly focus when an oscillation appears.
In our implementation, the encoder was trained for 30 iterations, followed by 100 iterations where the whole Transformer was trained. 
This was iterated 10 times.

\subsection{Genetic Algorithm implementation}

GA implementation: 
The platform consists of five rows with five motors.
Rows one and five were left for I/O, therefore the GA will optimize rows two, three and four, for a total of 15 motors.
In our GA implementation, each motor was considered a gene, therefore the genome was an array with 15 elements.
Our GA followed the standard approach with the following operations: ranking, selection using the roulette algorithm, crossover (two point) and mutation.
Our GA used a population size of 512, with 50 elite individuals, 100 generations and a mutation rate of 0.05.

\subsection{Reinforcement Learning and controller implementation}

RL controller: It was a single layer of 1024 neurons, which was connected to an output layer with 25 neurons - equal to the number of motors, using a sigmoid activation function. 
Given a sequence of motors and a sequence describing how the experiment changes, the first step was to feed-forward these two sequences through the Transformer model with the new controller output, to predict a new sequence of motors which would manipulate the chemistry towards a user defined target.
The loss function of the controller received three variables: (1) the sequence of motors just described, (2) the sequence describing the experiment as just described and (3) the motor configuration prediction just outputted by the controller.
The loss function would then concatenate the new motor prediction (3) to the old sequence of motors (1), removing the oldest element to keep a sequence of constant length -- which in our case was 150.
Then we would input this new motors sequence (1+3) with the sequence describing the experiment (2) through the Transformer described in section \ref{transarch} to predict a new chemical state.
Based on this chemical state, we would consider if the new motor configuration (3) had a positive or a negative effect against the user defined target function, and then reward it accordingly.
Using this loss function the controller layer was trained using backpropagation.

\subsection{Increasing the size of the IO feature space}

The Transformer model described in section \ref{transarch} aims to map the volution of 25 (5 by 5) input motors to 25 (5 by 5) chemical cells.
Here we aimed to increase the feature space, by mapping the evolution of $N^2$ (N by N) input motors to $N^2$ (N by N) chemical cells.
To do so, the transformer model was used as a kernel that convolved through the new input space (which had a shape of N by N by sequence).
The kernel size was the input size of the trained Transformer, that is 5 by 5, and the stride size was set to 1.
The output of each kernel operation was the prediction of the Transformer model.
Out of this prediction, only the centre cell was kept.
Following this, a target feature space of size $N^2$ would require $N^2$ transformer predictions.

\section*{Source code availability}

Implementation details and source code can be found at: https://github.com/thephet/SciTransformer

\section*{Acknowledgment}

We would like to thank Prof. Leroy Cronin for sharing the dataset used in this research.

\bibliographystyle{IEEEtran}
\bibliography{library}

\end{document}